\documentclass{article}
\PassOptionsToPackage{square,numbers}{natbib}
\usepackage[final]{neurips_2023}
\usepackage{amssymb}
\usepackage{amsmath,amsfonts}
\usepackage{algorithmic}
\usepackage{array}
\usepackage[caption=false,font=normalsize,labelfont=sf,textfont=sf]{subfig}
\usepackage{textcomp}
\usepackage{stfloats}
\usepackage{url}
\usepackage{verbatim}
\usepackage{graphicx}

\usepackage{comment}
\usepackage{array}
\usepackage[utf8]{inputenc} 
\usepackage[T1]{fontenc}    
\usepackage{hyperref}       
\usepackage{url}            
\usepackage{booktabs}       
\usepackage{amsfonts}       
\usepackage{nicefrac}       
\usepackage{microtype}      
\usepackage{lipsum}
\usepackage{fancyhdr}       
\usepackage{graphicx}       
\graphicspath{{media/}}     
\usepackage{tabularx}
\usepackage{makecell}
\usepackage{multirow}
\usepackage{siunitx}
\usepackage{caption}
\usepackage{subfig}
\usepackage{changepage}

\title{On-orbit model training for satellite imagery with label proportions}

\author{%
  Raúl Ramos-Pollán \\
  Universidad de Antioquia\\
  Calle 70, N 52-21\\
  Medellín, Colombia\\
  \texttt{raul.ramos@udea.edu.co}\\
  \em{Corresponding author}\\
  \And
  Fabio A. González\\
  Universidad Nacional de Colombia\\
  Bogotá, Colombia\\
  \texttt{fagonzalezo@unal.edu.co}
}

\begin{document}
\maketitle


\begin{abstract}
This work addresses the challenge of training supervised machine or deep learning models on orbiting platforms where we are generally constrained by limited on-board hardware capabilities and restricted uplink bandwidths to upload. We aim at enabling orbiting spacecrafts to (1) continuously train a lightweight model as it acquires imagery; and (2) receive new labels while on orbit to refine or even change the predictive task being trained. For this, we consider chip level regression tasks (i.e. predicting the vegetation percentage of a 20 km$^2$ patch) when we only have coarser label proportions, such as municipality level vegetation statistics (a municipality containing several patches). Such labels proportions have the additional advantage that usually come in tabular data and are widely available in many regions of the world and application areas. This can be framed as a Learning from Label Proportions (LLP) problem setup. LLP applied to Earth Observation (EO) data is still an emerging field and performing comparative studies in applied scenarios remains a challenge due to the lack of standardized datasets. In this work, first, we show how very simple deep learning and probabilistic methods (with {\raise.17ex\hbox{$\scriptstyle\sim$}}5K parameters) generally perform better than standard more complex ones, providing a surprising level of finer grained spatial detail when trained with much coarser label proportions. Second, we publish a set of benchmarking datasets enabling comparative LLP applied to EO, providing both fine grained labels and aggregated data according to existing administrative divisions. Finally, we show how this approach fits an on-orbit training scenario by reducing vastly both the amount of computing and the size of the labels sets. Source code is available at \href{https://github.com/rramosp/llpeo}{\texttt{https://github.com/rramosp/llpeo}}.

\textbf{keywords} earth observations, remote sensing, learning with label proportions, on orbit AI, deep learning

\end{abstract}

\section{Introduction}

Earth Observation (EO) has had remarkable progress with the rise of deep learning (DL) methods in the last years and this potential is fueled by the ever increasing  availability of satellite imagery \cite{zhu2017deep}. According to the UCS Satellite Database\footnote{https://www.ucsusa.org/resources/satellite-database}, as of May 2022 there were 470 optical satellites in orbit, 102 radar satellites and 62 tagged as producing some sort of imaging (hyperspectral, multispectral), among others. Only in ESA's Sentinel missions, 80 PB of user-level data were downloaded during 2021 \cite{copernicus2021}.

EO satellites acquire data and downlink it in raw or lightly processed or compressed formats for further processing on the ground. Time-sensitive applications  such as emergency response or near real time change detection might be hindered by latencies in these processing pipelines and, thus, shifting part of them to  on-board platforms starts to make sense in specific use cases. This is likely to be increasingly the case as the number of EO satellites in orbit grows, with further restrictions on the radio-frequency spectrum and licensing availability \cite{selva2012survey}. Due to this, there is an emerging interest in on-orbit (or on-board, or on-edge) machine learning capabilities \cite{kothari2020final} and its applications \cite{ruuvzivcka2022ravaen} \cite{mateo2021towards} \cite{mateo2023orbit} .

In this work we aim at fully enabling machine learning pipelines on-orbit, from training to inference, including the possibility of dynamically updating the predictive task that is trained on-board. This is generally constrained by the limited on-board hardware capabilities to train models and restricted uplink bandwidths to upload new labels. Assuming that we use the input imagery as acquired by the orbiting platform as input data for model training or inference, we are therefore bound to train very lightweight models and consider only label sets with very small footprints.

We approach these challenges by framing them as a Learning from Label Proportions (LLP) problem, where models are trained to produce fine grained predictions when only coarse label proportions are available at training time. We build upon the observation tat the geospatial nature of EO data creates particular conditions where LLP approaches can enable on-orbit machine learning pipelines. Specifically, we exploit the fact that aggregated information is generally available on many topics (demographics, land use, industrial activity, etc.) at some administrative unit level that we denote as \emph{communes} (or \emph{municipalities}). Images are grouped in  bags, where each bag holds all images of the same commune, and only the proportions of labels in each bag are available for training. Our intuition is that the distribution of the resulting image bags is likely to be conditioned by the specific geophysics and demographics of that commune, probably inducing significant correlation among bag members simply because they belong to similar geographies. This way, the footprint of commune level label proportions (storage size) is orders of magnitude smaller than fine grained labels reducing both the labels footprint and computing requirements for models.

Our models produce image level predictions on satellite imagery when only label proportions at a coarser spatial geometry (communes) are available. We use the term \emph{chip} to denote satellite imagery that has been chopped into a grid of smaller images (chips) of equal size. Given an input chip (100 $\times$ 100 RGB pixels in our case) we want to (1) have models predicting chip-level proportions of classes; (2) build loss functions that only use label proportions at the commune level. Models may also produce segmentation maps, although we do not explicitly address it in this work.

The contributions of this work are the following. First, we show how simple deep learning and probabilistic methods generally perform better than standard more complex ones, providing a surprising level of finer grained spatial detail when trained with much coarser label proportions. Second, since EO for LLP is an emerging field and there are virtually no datasets available to perform comparative, reproducible studies, we provide a set of benchmarking datasets enabling comparative LLP applied to EO, providing both fine grained labels and aggregated data according to existing administrative divisions. Finally, we argue how both aspects combined might enable on-orbit machine learning pipelines.

This paper is structured as follows. The following sections discusses previous works. In Section \ref{sec:data} we describe the datasets we produced in this study and that we make publicly available. Section  \ref{sec:methods} details the models and experimental setups we used. Section \ref{sec:results} describes and illustrates the results obtained and Section \ref{sec:onorbit} discusses considerations for on-orbit training and inference arising from this particular setting and results. Finally Section \ref{sec:conclusions} draws some conclusions.

\section{Previous works}
\label{sec:prevwork}
It is well known that DL methods are both computationally expensive and overwhelmingly hungry for labeled data. Already on ground, this endemic scarcity of labeled data is one of the main hurdles to effectively exploit DL methods in EO \cite{wang2022self}.  A number of approaches have been developed to address the labeled data scarcity challenge in EO, including a variety of methods under self supervised learning \cite{wang2022self}, weakly supervised methods \cite{fasana2022weakly} or, still timidly, foundation models \cite{lacoste2021toward}. Weakly supervised methods attempt to exploit low quality labels which are likely easier or less costly to gather than annotations for supervised learning. These labels might be incomplete (missing), innacurate (noisy) or inexact (coarse-grained). 

Learning from label proportions (LLP) \cite{hernandez2019framework} is one such approaches, in which full fledged labels are not directly available but only some summary of the prorportions of the classes one would like to predict. In a typical LLP setting in image classification, images are grouped in bags and only the proportion of classes in the bags are known. Most approaches assume certain random distribution of images in bags and measure how different random distributions affect models' outcomes \cite{tsai2020learning} \cite{la2022weak}. 
There is already some work on using DL with LLP \cite{shi2020deep} and, very incipient, in EO \cite{la2022learning} \cite{la2022learningCO} \cite{ding2017learning}.

\section{Data}
\label{sec:data}

We use Sentinel-2 RGB imagery as input data for our models at a resolution of 10m. For targets we use label proportions derived from (1)
 the ESA World Cover 2020 land use / land cover at a resolution of 10m, denoted by \texttt{esaworldcover} and (2) EU Joint Research Cent
er population density estimate from their Global Human Layer settlements datasets at a resolution of 250m, denoted by \texttt{humanpop}. These three data sources are available from Google Earth Engine (GEE) and described in the sections below. As part of this work, we developed  the toolkit \texttt{geetiles}\footnote{\href{https://github.com/rramosp/geetiles}{https://github.com/rramosp/geetiles}} to download and create aligned image chips for each data source, and compute label proportions at different spatial granularity levels.

We focused on two different regions in terms of land features and demographics: (1) Belgium, the Netherlands and Luxembourg (\texttt{benelux}), and (2) a region in the north west of Colombia (\texttt{colombia-ne}). Figures \ref{fig:aoi-benelux} and \ref{fig:aoi-colombia} detail these two areas of interest. Notice that each chip is always associated with the \emph{commune} (municipality or administrative division) to which it belongs. 

We therefore created four datasets as shown in Table \ref{tab:datasets} with the distribution of classes as shown in Figure \ref{fig:class-distribution}. Note that \texttt{humanpop} on \texttt{colombia-ne} is overwhelmingly dominated by class 0 (mostly unpopulated areas) by $\sim$99\%. Chips have a size size 100 $\times$ 100 pixels so that for convenience each chip covers 1 km$^2$, with a nominal 10m/pixel resolution, this is, if the original dataset resolution is lower, it is upsampled using nearest neighbours. Each dataset contains as many chips as km$^2$ covered, and each chip is comprised of the \texttt{sentinel2-rgb} image, the \texttt{esa-world-cover} or the \texttt{humanpop} segmentation map and the corresponding label proportions at chip level and at commune level (see Section \ref{sec:label-proportions}). Observe that all chips within the same commune contain the same commune-level label proportions, which are the ones obtained by aggregating the proportions of all chips in that commune.

\begin{table*}
\begin{center}
\begin{tabular}
            {>{\arraybackslash}p{2.5cm}
             >{\centering\arraybackslash}p{2.7cm}
             >{\centering\arraybackslash}p{6cm}
            }

\toprule
labels / region &original resolution of labels & land surface / available at\\
\midrule
\texttt{colombia-ne}  & & 69193 km$^2$\\
\;\;\;\;\small\texttt{esaworldcover}  & 10m &  \small https://zenodo.org/record/7935303  \\
\;\;\;\;\small\texttt{humanpop}   & 250m & \small https://zenodo.org/record/7939365\\
\hline
\texttt{benelux} & &72213 km$^2$\\
\;\;\;\;\small\texttt{esaworldcover} & 10m & \small https://zenodo.org/record/7935237\\
\;\;\;\;\small\texttt{humanpop}   & 250m & \small https://zenodo.org/record/7939348\\
\bottomrule
\\

\end{tabular}
\caption{Datasets generated in this work. The Sentinel 2 image chips are the same in both \texttt{colombia-ne} datasets, they differ on the labels. Likewise is the case with both \texttt{benelux} datasets. Observe that we train our models with label proportions that we obtain from these labels at coarser geometries (communes or municipalities). We only use the actual finer grained labels at to compute chip level performance metrics. In a real world scenario these fine grained labels would \emph{not} be available, only the label proportions.
\label{tab:datasets} 
}
\end{center}
\end{table*}

Note that in all cases, we generated datasets with chip level segmentation labels as fine grained as possible, so that we can group them at desired granularity levels for training and measuring performance (see Section \ref{sec:label-proportions} below). Also, we grouped and discretized the original classes in \texttt{esaworldcover} and \texttt{humanpop} provided by GEE as described in the follow
ing sections.

\subsection{Input imagery from Sentinel 2 RGB}
As input data for our models we use RGB imagery from Sentinel 2 at 10m resolution. Tiles are extracted from the Copernicus S2 SR Harmonized  GEE 
 dataset \cite{esas2}, clipping the original data range at 3000 and standardizing to the range [0,1]. For each pixel we take the cloudless time series of 2020 and compute the median, see our source code at \href{https://github.com/rramosp/geetiles/blob/main/geetiles/defs/sentinel2rgbmedian2020.py}{\texttt{geetiles/defs/sentinel2rgbmedian2020.py}}.
 
\subsection{Task 1: Land use / Land cover}
Tiles for targets of this task are extracted from the ESA WorldCover 10m V100 2020 GEE dataset \cite{zanaga_daniele_2021_5571936}. Quoting documentation at GEE \emph{The European Space Agency (ESA) WorldCover 10 m 2020 product provides a global land cover map for 2020 at 10 m resolution based on Sentinel-1 and Sentinel-2 data. The WorldCover product comes with 11 land cover classes and has been generated in the framework of the ESA WorldCover project, part of the 5th Earth Observation Envelope Programme (EOEP-5) of the European Space Agency. }

The original labels are grouped and mapped according to the Table \ref{table:esaworldcover-classes} so that there is a more even class distribution. The resulting distribution can be seen in Figure \ref{fig:class-distribution}. See our source code at \href{https://github.com/rramosp/geetiles/blob/main/geetiles/defs/esaworldcover.py}{\texttt{geetiles/defs/esaworldcover.py}}.

\begin{table}
\begin{center}

\begin{tabular}
            {>{\centering\arraybackslash}p{1.5cm}
                >{\centering\arraybackslash}p{1.8cm}
                >{\centering\arraybackslash}p{3cm}
                >{\raggedright\arraybackslash}p{5cm}
            }

\toprule
class id &  class\newline description & class id in source GEE dataset &  class descriptions in source GEE dataset aggregated\\
\midrule
0             &  Infrequent             & 70 \newline 80 \newline 90\newline 95\newline 100\newline & Snow and ice \newline Permanent water bodies\newline Herbaceous wetland \newline Mangroves \newline Moss and lichen\\

1 & Treecover   &10 \newline &Treecover\newline \\

2 & Vegetation & 20 \newline 30 \newline 60 \newline & Shrubland \newline Grassland \newline Bare / sparse vegetation\\

3 & Cropland   & 40 \newline &Cropland\newline \\

4 & Built-up   &50 \newline &Built-up\newline \\
\bottomrule
\\

\end{tabular}
\caption{Mapping of \texttt{esaworldcover} classes.
\label{table:esaworldcover-classes}
}
\end{center}
\end{table}

\subsection{Task 2: Population density estimation}

Tiles for target prediction are extracted from the Global Human Settlement Layers, Population Grid 1975-1990-2000-2015 GEE dataset \cite{ghsl}. Quoting the documentation at GEE. \emph{The GHSL relies on the design and implementation of new spatial data mining technologies allowing to automatically process and extract analytics and knowledge from large amount of heterogeneous data including: global, fine-scale satellite image data streams, census data, and crowd sources or volunteered geographic information sources. This dataset depicts the distribution and density of population, expressed as the number of people per cell, for reference epochs: 1975, 1990, 2000, 2015. Residential population estimates were provided by CIESIN GPW v4. These were disaggregated from census or administrative units to grid cells, informed by the distribution and density of built-up as mapped in the GHSL global layer per corresponding epoch.
}

\begin{table}
\begin{center}
\begin{tabular}
            {>{\centering\arraybackslash}p{3cm}
                >{\arraybackslash}p{6cm}
            }

\toprule
class number &  population density interval \\
\midrule
0             & 0 to 15 inhabitants/km$^2$\\

1 & 16 to 1600 inhabitants/km$^2$ \\

2 & more than 1600 inhabitants/km$^2$ \\

\bottomrule
\\

\end{tabular}
\caption{Mapping of \texttt{humanpop} classes.
\label{table:humanpop-classes}
}
\end{center}
\end{table}

We take the band \texttt{population\_count} of 2015 originally in the range (0 .. 1.3$\times 10^6$), which represents the number of inhabitants in a 250m $\times$ 250m square, and discretize it into 3 classes as described in Table \ref{table:humanpop-classes}. See our source code at 
\href{https://github.com/rramosp/geetiles/blob/main/geetiles/defs/humanpop2015.py}{\texttt{geetiles/defs/humanpop2015.py}}. Since the original resolution of this dataset is 250m GEE returns the tiles upsampled to match the 10m resolution used in Sentinel2-RGB and \texttt{esaworldcover}. See Figure \ref{fig:image_samples}.

\subsection{Label proportions}
\label{sec:label-proportions}
\begin{figure*}[h]
\caption{\texttt{benelux} selected RGB images from the commune of Ichtegem (Belgium) with fine grained chip level labels (rows two and four) and proportions at chip level (rows 3 and 5). The commune level proportions are shown besides the chip level proportions for \texttt{esaworldcover} and \texttt{humanpop}. Recall that we are training using this \emph{commune proportions} shown here assuming that chips do not have individual labels. When a chip intercepts more than one commune, such as chip $\texttt{006c47afde9e5}$ below, its associated commune level proportions are obtained by combining the proportions of the communes it overlaps, weighted proportionally to the amount of overlapping with each commune. Proportions and labels for individual chips are used only to compute performance metrics. See Tables \ref{table:esaworldcover-classes} and \ref{table:humanpop-classes} for label meanings.}

\centering
\includegraphics[width=1.0\textwidth]{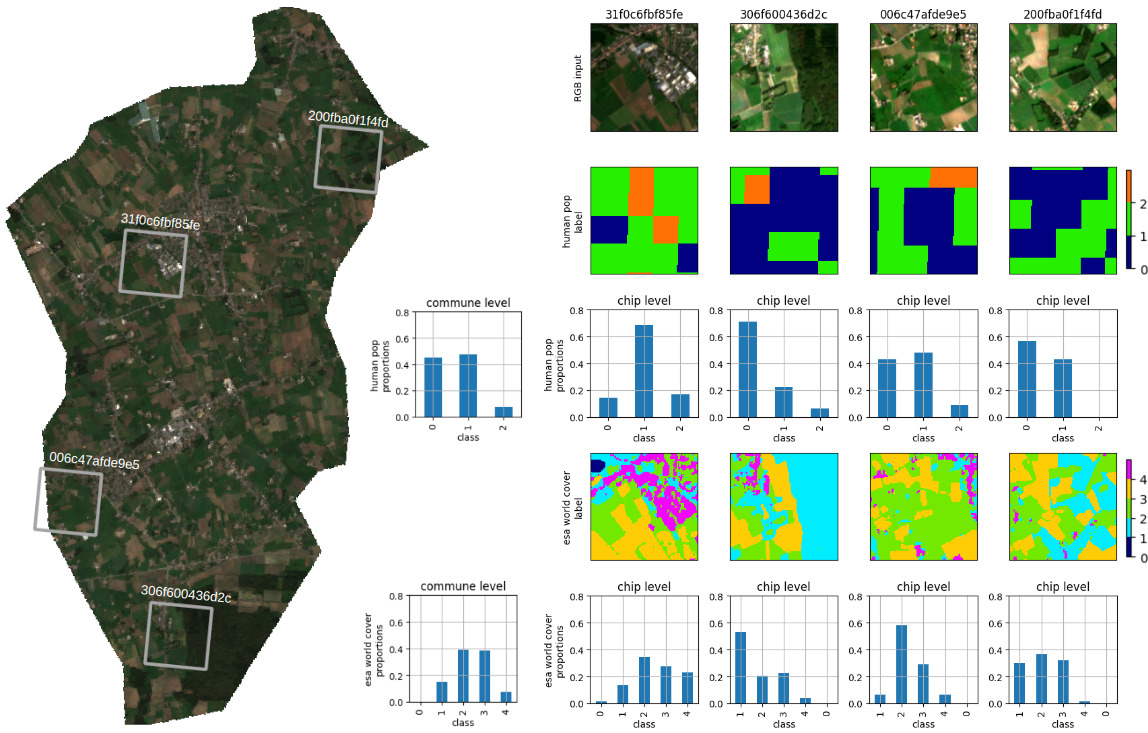}
\label{fig:image_samples}
\end{figure*}

Both from \texttt{esaworldcover} and from \texttt{humanppop}, we obtain label proportions at the chip level and \emph{commune} level. These correspond to the bar charts in Figure \ref{fig:image_samples}. \emph{Commune} level proportions are much coarser but typically available in many cases. We train with these \emph{commune} level proportions. Chip level proportions, which are finer grained but more scarce (as they are summary of the chip pixels) are only used for validation in this work.

\begin{figure*}[h]
\caption{\texttt{benelux} area of interest, covering 72.213 km$^2$ (top left). Tiling of $1km \times 1km$ over the surroundings of Amsterdam (bottom left). Sub
division in communes (municipalities) used to compute label proportions (right). }
\centering
\includegraphics[width=1.0\textwidth]{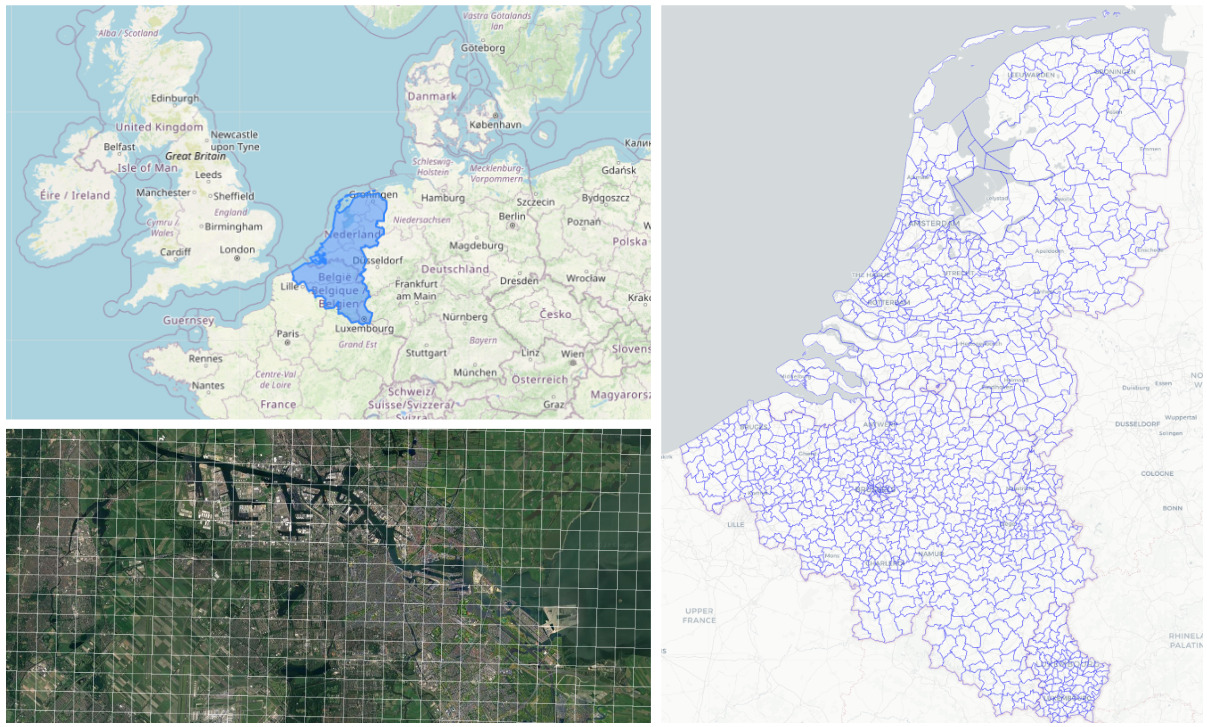}
\label{fig:aoi-benelux}
\end{figure*}

\begin{figure*}[h]
\caption{\texttt{colombia-ne} area of interest, covering 69.191 km$^2$ (top left, on the north west of South America). Tiling of $1km \times 1km$ over the surroundings of the city of Bucaramanga (bottom left). Subdivision in communes (municipalities) used to compute label proportions (right). }
\centering
\includegraphics[width=1.0\textwidth]{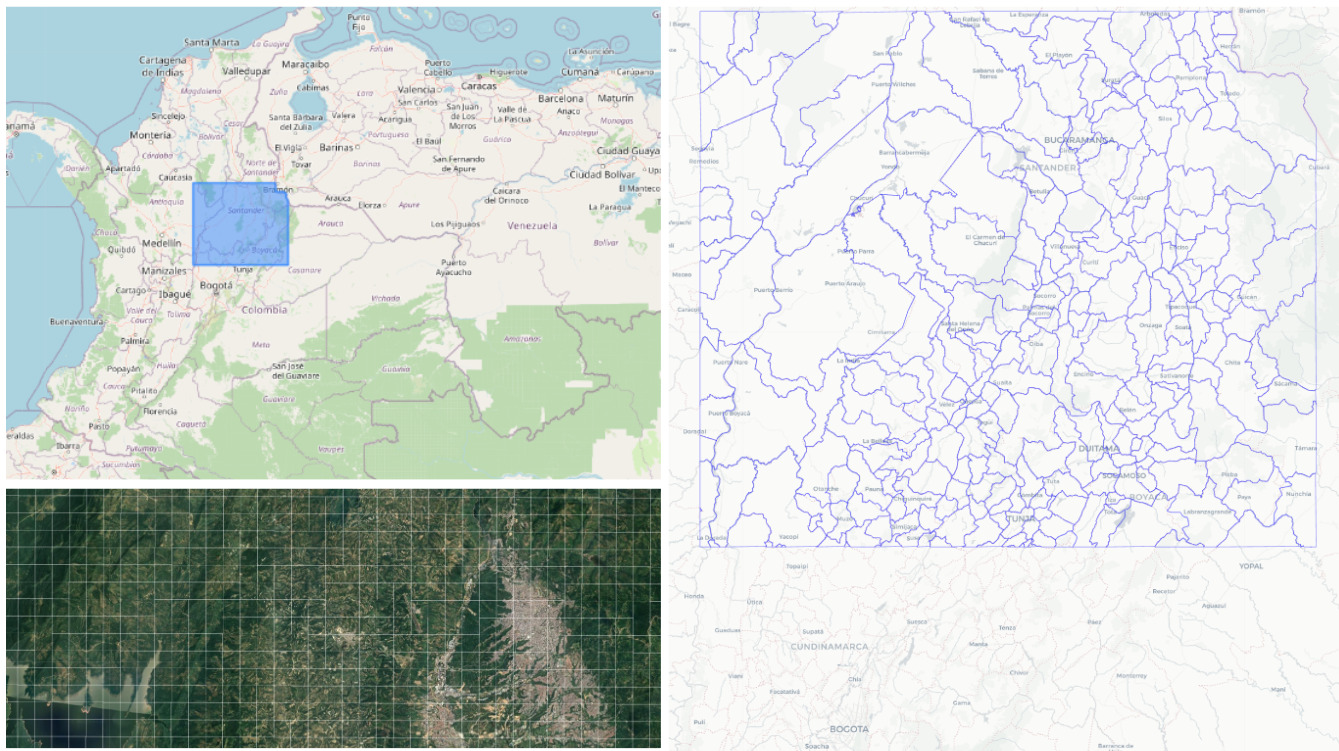}
\label{fig:aoi-colombia}
\end{figure*}

\emph{Communes} roughly correspond to the notion of \emph{municipality} in most countries. Commune definitions for \texttt{benelux} are taken from EUROSTAT Reference Data \cite{eurostat_communes}, and for \texttt{colombia-ne} they are taken from National Reference Framework of DANE (Departamento Nacional de Estadística) \cite{dane_mgn}.  See Figures \ref{fig:aoi-benelux} and \ref{fig:aoi-colombia} for their geographical extent and communes definition.

\subsubsection{Data splitting}
\label{sec:data-splitting}
We do the customary train / validation /  test split of the chips in each dataset, using validation for model selection and test for reporting metrics. We want to minimize the data leakage regularly present in geospatial datasets due to the contiguous chips being in different splits. For this, we split the chips in geographical bands at 45° (nw to se) and we alternate assignment between train, validation and test.

\begin{figure*}[h]
\caption{Data splitting for \texttt{benelux} (left) and \texttt{colombia-ne} (right) so that any commune (municipality) has all its chips within the same split. Train is purple, yellow is test, green is validation.}
\centering
\includegraphics[width=1.0\textwidth]{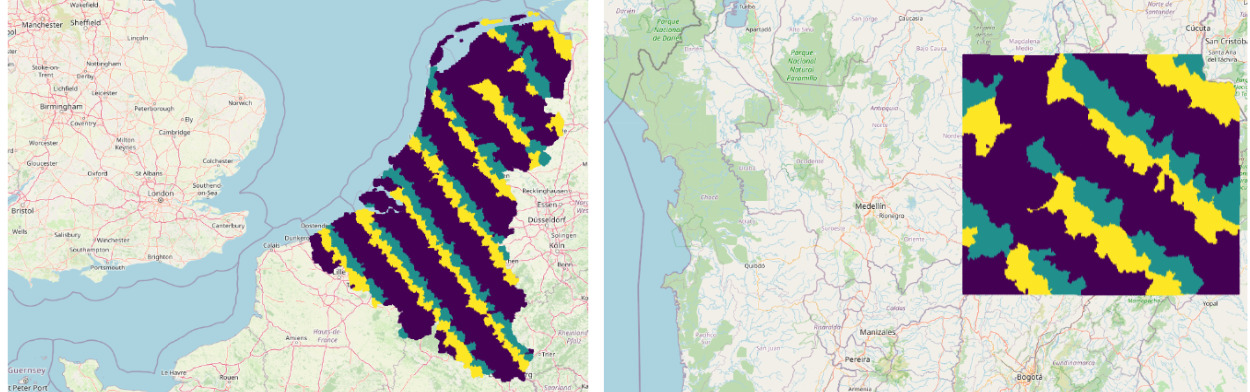}
\label{fig:datasplits}
\end{figure*}

Furthermore, recall that we are training with label proportions at \emph{commune} level, and all chips within the same commune will be trained against the same label proportions target. Hence, we also want to avoid leakage within communes and, thus, we modify the initial assignment at 45° geographical bands so that all chips within the same commune are in the same split. Figure \ref{fig:datasplits} shows this split for both areas of interest. This split is also included in the datasets we make available.

In all, the \texttt{benelux} dataset contains 43092 chips for train, 15011 for test and 14110 for validation. And the dataset for \texttt{colombia-ne} contains 34715 chips for train, 17506 for test and 16972 for validation.

\section{Methods}
\label{sec:methods}

Our aim is to develop models that take as input an RGB 100$\times$100 image and produce a vector of $n$ components with the proportions of the occurrence of the $n$ classes in that chip. In the case of \texttt{esaworldcover} we deal with 5 classes, and in the case of \texttt{humanpop} we deal with 3 classes. 

We insist that we train using only commune level proportions, although in this exercise we do have chip level segmentation labels. These are used only for measuring performance at the chip level. Thus, we are measuring performance at a finer grain level that what we train with. In a real world scenario, where only tabular is availale, even measuring performance at this finer detail would not be possible.

As a subproduct we also want models to produce a segmentation map at some resolution (possibly coarser than the input image) indicating where those classes occur. Since we only have tabular data of label proportions at a communes level so we can only build loss functions with the label proportions output of our models, but not with the output segmentation map, should a method produce it. In our datasets, we do have fine grained segmentation labels (and thus, chip level proportions) but we use these only to measure methods performance.

A straightforward way of using existing segmentation methods is to compute directly the resulting label proportions from the segmentation output and then use a differentiable loss with the commune level proportions. In this setting the loss for all chips within the same commune will be built against the same label proportions.

In all models we used the mean squared error (MSE) loss comparing the vector of label proportions produced by a model when presented with a single input chip, with the communes level label proportions of the commune containing the input chip. Again, even if all models described below produce segmentation maps, these are not used in any loss function, only for measuring performance later on.

\subsection{Models}
\begin{figure}[h]
\caption{Simple convolutional model used to predict label proportions.}
\centering
\includegraphics[width=0.8\textwidth]{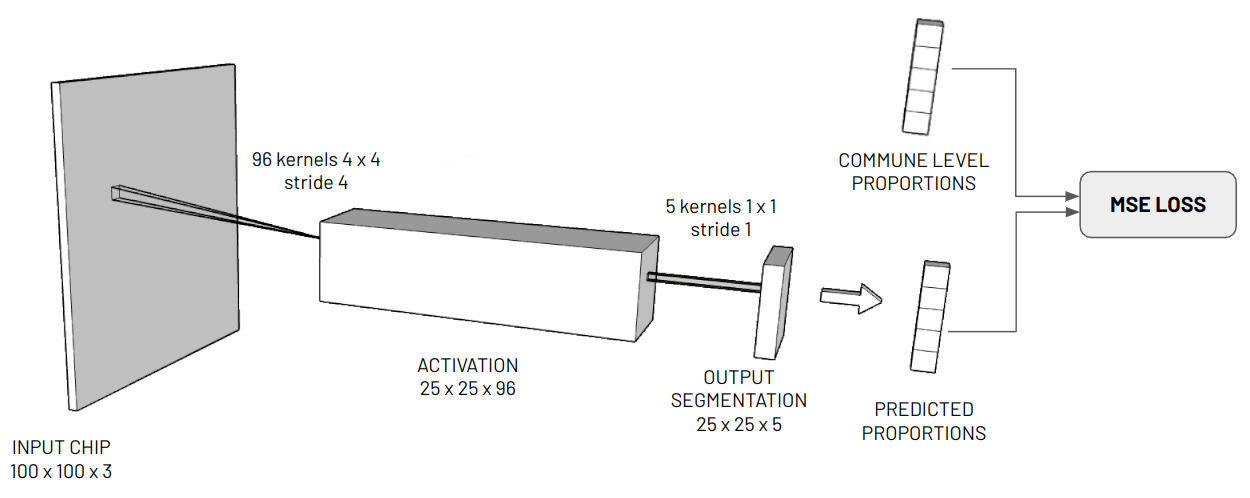}
\label{fig:downsampling}
\end{figure}

\paragraph{Simple downsampling convolutions (\texttt{downconv})} As shown in Figure \ref{fig:downsampling} this is a very simple model w
ith two convolutional layers. A first layer with non overlapping 4$\times$4 convolutions produces activations maps of dimension 25 $\times$ 25; and a second layer with 1$\times$1 convolutions so that there is one output channel per class keeping the 25 $\times$ 25 dimension. We include a softmax output so that we have a probability distribution at each output cell over the label classes and, finally, we compute the overall label proportions from this output by summing up the probabilities of each output channel and normalizing to add up to 1. This model has 5189 parameters.

\paragraph{Probabilistic model with quantum kernel mixtures (\texttt{qkm})} We used the probabilistic framework described in detail here \cite{gonzalez2023quantum}, which originated the ideas shown in \cite{Gonzalez2022LearningFeatures} . In short, from an input chip we generate a bag of small sized patches, and set up a classification task for each patch. The framework then builds a joint probability distribution $p(\textbf{x}, \textbf{y})$ based on a mixture of simpler distributions specified through function kernels, which is fully differentiable. Once the joint distribution is built, the framework is flexible enough to allow efficient computation of any statistical inference upon $p(\textbf{x}, \textbf{y})$. 

\begin{figure}[h]
\caption{Probabilistic kernel mixtures model.}
\centering
\includegraphics[width=0.8\textwidth]{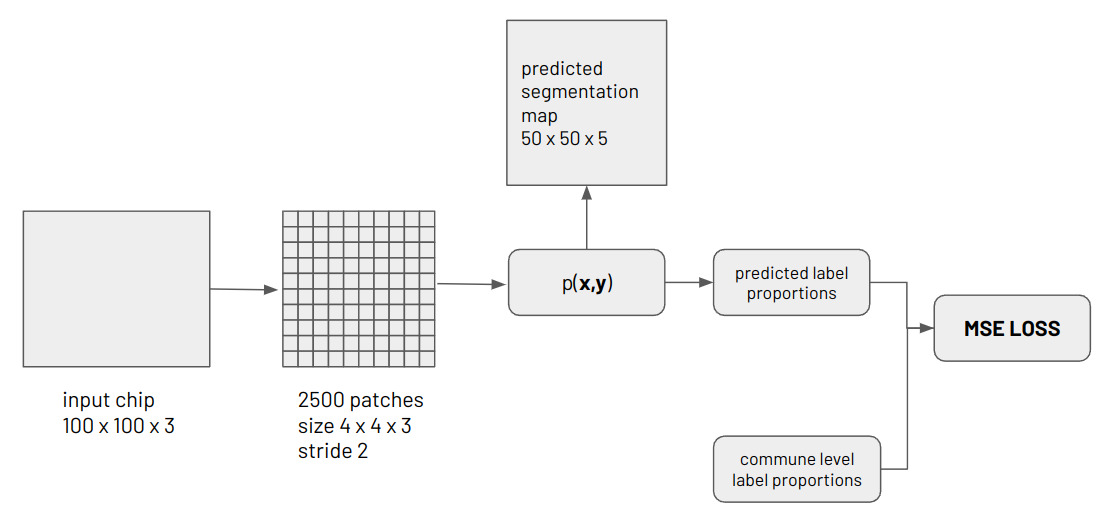}
\label{fig:qkm}
\end{figure}

Specifically, once trained, given the patches obtained by an input chip we can obtain $p(\textbf{y}|\textbf{x})$, a distribution over the output classes, which can then be matched against the corresponding commune level label proportions for regular training with a loss function. See Figure \ref{fig:qkm}. The method can also produce segmentation maps but, unlike the other methods we use here, the segmentation maps are not used to generate chip level label proportions (and thus they do not contribute to the loss function), they are simply another output.

In our experiments we use a patch size of 4$\times$4 with stride 2 and 64 kernel components resulting in a model with 3457 parameters. The predicted segmentation maps have a resolution of 50$\times$50 pixels.

\paragraph{Unet VGG 16 (\texttt{vgg16unet})} The standard Unet VGG16 architecture based on the \texttt{segmentation models} github package \cite{Yakubovskiy:2019} with 23.7M parameters. Given the network output segmentation map with size 100$\times$100$\times n$ (with $n$ classes) we compute the label proportions as described above with the model \texttt{downconv}.

\paragraph{Custom Unet (\texttt{customunet})} A custom Unet architecture with four convolutional layers and four upsampling layers with the customary skipped connections. This model has 482K parameters. Given the network output segmentation map with size 100$\times$100$\times n$ we compute the label proportions as described above with the model \texttt{downconv}.

\subsection{Metrics}
We measure the performance of the models at the chip level by using the chip segmentation map obtained by the GEE datasets and the label proportions computed from them. Note that this information will not be available in a real world scenario (only the tabular commune level label proportions as used for training), and we use it here to measure model performance, providing a crispier understanding of models' behaviour. This is also the reason behind our choice of the \texttt{esaworldcover} and \texttt{humanpop} datasets, since they provide chip level labels, even if it is at different resolutions.

\paragraph{Mean Absolute Error (MAE)} We compare predictions of label proportions using MAE, as it provides a simple metric easy to explain to non-experts. Since we are predicting proportions MAE has a very direct interpretation. A MAE of 8\% in a specific chip of 1 km$^2$ means that the average error across the classes in consideration represents 0.08 km$^2$.

\paragraph{F1 score} We compare the predicted segmentation maps with the original dataset labels with the F1 score at the pixel level. If required, we upsample predictions and labels to the chip size by using nearest neighbour interpolation. Observe that, in this case, for each pixel we select the class with highest predicted probability and then we compare it with the target to compute the score. This is a bit different from the way we compute label proportions from the same segmentation map to build the loss since, in that case, we do not \texttt{argmax} at the pixel level, but average out all probabilities. 

\subsection{Experimental setup}
We used the data splits described in Subsection \ref{sec:data-splitting}, using validation data for model selection and test data for performance reporting. Model selection was done using MAE at chip level on test data. We trained all models during 50 epochs with no early stopping, since we observed that, in many occasions, loss would stall during training for a few epochs to later on continue its descent. Still, we report the performance at the epoch which obtained the lowest validation loss so that we avoid reporting under estimations of performance when there is large overfitting and performance in validation degrades at the end of training.

\begin{figure*}[h]
\caption{Aggregated distributions of class proportions for 2*5*1082                                                             \texttt{colombia-ne} and \texttt{benelux} (a,b,c,d). Observe that the distributions are quite similar when aggregated from communes and from chips, as it should be, since the chips cover 100\% of communes. This is also a sanity check for the datasets. The small differences come from chips overlapping several communes at their borders. See Tables \ref{table:esaworldcover-classes} and \ref{table:humanpop-classes} for label meanings. Figure e) shows the distribution  of commune sizes in both AOIs. Since each chip is 1 $km^2$ this also represents the distribution of communes sizes in $km^2$.}
\centering
\includegraphics[width=1.0\textwidth]{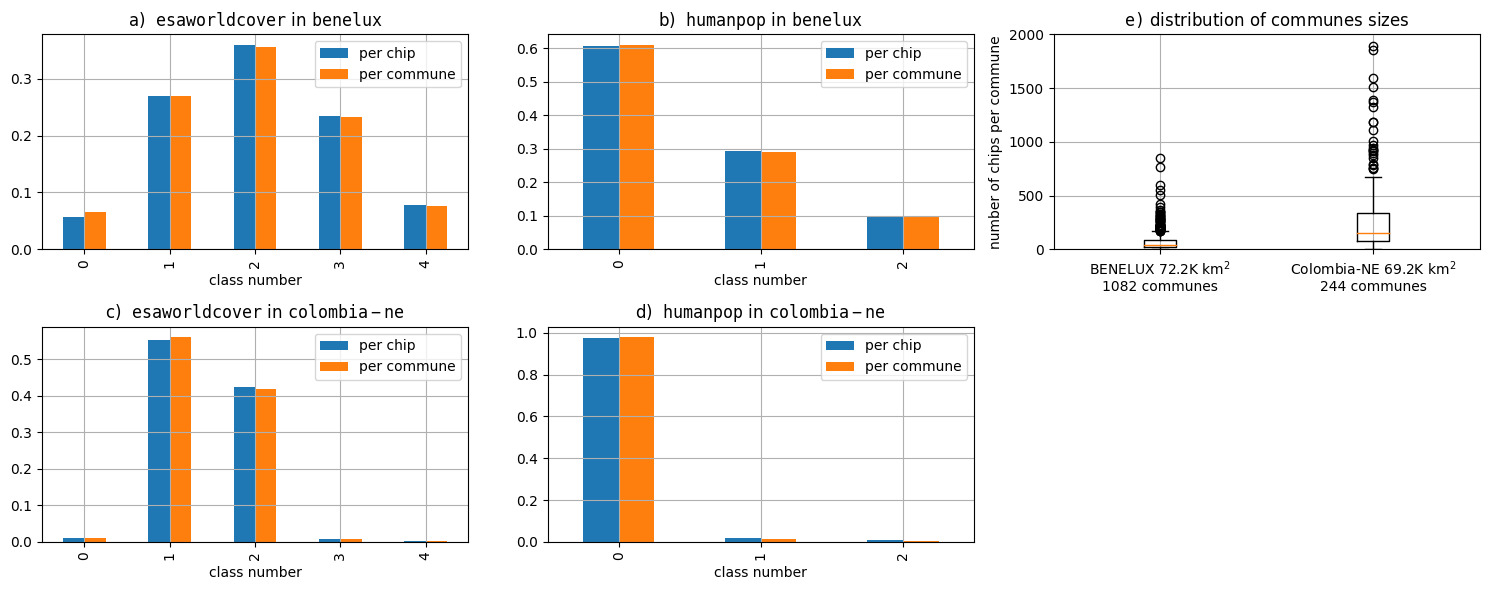}
\label{fig:class-distribution}

\end{figure*}

We did not perform experiments on \texttt{colombia-ne} for \texttt{humanprops} due to the huge class imbalance shown in Figure \ref{fig:class-distribution}. However this dataset is also made available as referenced in Table \ref{tab:datasets}.

For each model, except \texttt{vgg16unet}, and dataset we performed a hyperparameter search on their architectures. In the case of \texttt{downconv} and \texttt{qkm} we explored even patch sizes between 2 and 16 with strides 2, 4, 6 and the size of the patch whenever that made sense (stride less or equal than the patch size). For \texttt{downconv} we also explored number of filters between 8 and 96 and added up to 3 additional convolutional layers with stride 1 so that the output resolution was not further diminished. This resulted in models having from about 500 parameters up to 1M parameters. The search was not exhaustive (i.e. we did not run all possible combinations).

For \texttt{qkm} we also explored having 16, 32 and 128 components in the kernel mixture. This resulted in models having from 300 parameters to around 100K parameters. And for \texttt{customunet} we explored with different number of layers from 2 layers (25K parameters) to 5 layers (1.9M parameters). 

In \texttt{vgg16unet} we also tested using the pretrained weights from Imagenet but significant improvement was observed.  Other models available in \cite{Yakubovskiy:2019} were tried (ResNet, MobineNet, InceptionNet) but none worked better than \texttt{vgg16unet}.

Experiments were performed partially on Google Colab Pro and partially on a machine with an NVIDIA GeForce RTX 2080 with 8GB of GPU memory, 16GB of RAM running an Intel Core i7-8700 CPU.

\section{Results}
\label{sec:results}
Table \ref{tab:results} shows the best results for each model and dataset obtained in our experimentation. Observe how simpler models (i.e. with a small number of parameters) perform consistently better than larger models in both MAE and F1 when using the \texttt{esaworldcover} dataset regardless the geographical region. We consider this to be significant as both regions have very different geophysical features and class distributions. For \texttt{humanpop} on \texttt{benelux} there are not major differences in the metrics, specially in MAE. 


\begin{table*}
\caption{Overall experimental results reported on test data. We use the label proportions at the chip level only for producing these metrics (not for training). Given the pixel level labels for each chip we compute the label proportions and compare both the segmentation model output (F1) and its proportions (MAE). Observe the simple models proposed in this work (\texttt{qkm} and \texttt{downconv}) produce competitive performace as compared with larger segmentation models.
\label{tab:results}
}
\begin{center}

  \begin{tabular}{lc cc cc cc cc cc}
    \toprule
      \multicolumn{2}{c}{} &
      \multicolumn{2}{c}{\shortstack[c]{\texttt{benelux} }} &
      \multicolumn{2}{c}{\shortstack[c]{\texttt{colombia-ne} }}\\

      {Model} & {parameters} &  {MAE} & {F1} & {MAE} & {F1}  \\
      
      \midrule
      \texttt{esaworldcover} \\
      \midrule
    \texttt{downconv} (this work)& 5189   & \textbf{0.068} & \textbf{0.636}  & 0.059 & \textbf{0.808} \\
    \texttt{qkm} (this work)  & 3457   & \textbf{0.068} & 0.633 & \textbf{0.053} & 0.778 \\
    \texttt{vgg16unet}    &23.7M  & 0.085 & 0.548  & 0.074 & 0.676 \\
    \texttt{customunet}     &482K   & 0.104 & 0.425 & 0.082 & 0.613 \\
      \midrule
    \texttt{humanpop} \\
    
      \midrule
    \texttt{downconv} (this work)& 5189   & 0.139 & 0.593 & \textbf{0.015} & \textbf{0.978} \\
    \texttt{qkm} (this work)  & 3457   & \textbf{0.137} & 0.581 & \textbf{0.015} & \textbf{0.978} \\
    \texttt{vgg16unet}    &23.7M  &  0.139 & \textbf{0.624} & 0.021 & 0.967\\
    \texttt{customunet}     &482K & \textbf{0.137} & 0.550 & 0.026 & 0.958 \\
    \bottomrule
  
  \end{tabular}

\end{center}
\end{table*}

\begin{figure}[h]
\caption{Results for models on \texttt{esaworldcover} vs. training time. As reference, we include (1) the performance of a model doing regression to the mean (black line), which would be an upper limit for MAE; and (2) the performance of a vgg16 model trained with the fine grained segmentation maps at chip level with cross entropy loss at the pixel level (red line), which would be a lower limit for MAE and an upper limit for F1.}
\centering
\includegraphics[width=1.0\textwidth]{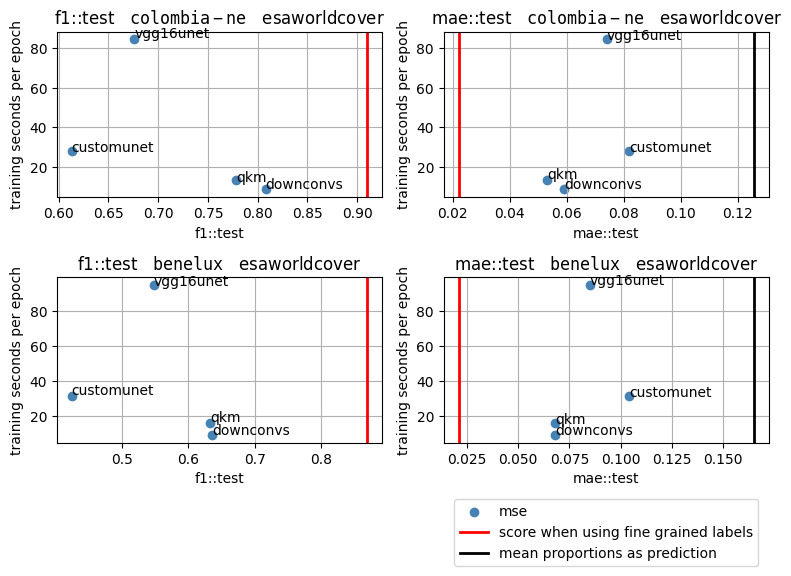}
\label{fig:esaworldcover-results-vs-time}
\end{figure}

\begin{figure}[h]
\caption{Results for models on \texttt{humanpop} vs. training time. As reference, we include (1) the performance of a model doing regression to the mean (black line), which would be an upper limit for MAE; and (2) the performance of a vgg16 model trained with the fine grained segmentation maps at chip level with cross entropy loss at the pixel level (red line), which would be a lower limit for MAE and an upper limit for F1. Observe that, in the case of \texttt{colombia-ne} all results lay within a very narrow margin, indicative of a regression to the mean performance overall, fluctuations allowing. Even, considering the F1 score, there is no difference between training with label proportions and the fine grained per-chip segmentation labels (F1 results falling on the red line).}
\centering
\includegraphics[width=1.0\textwidth]{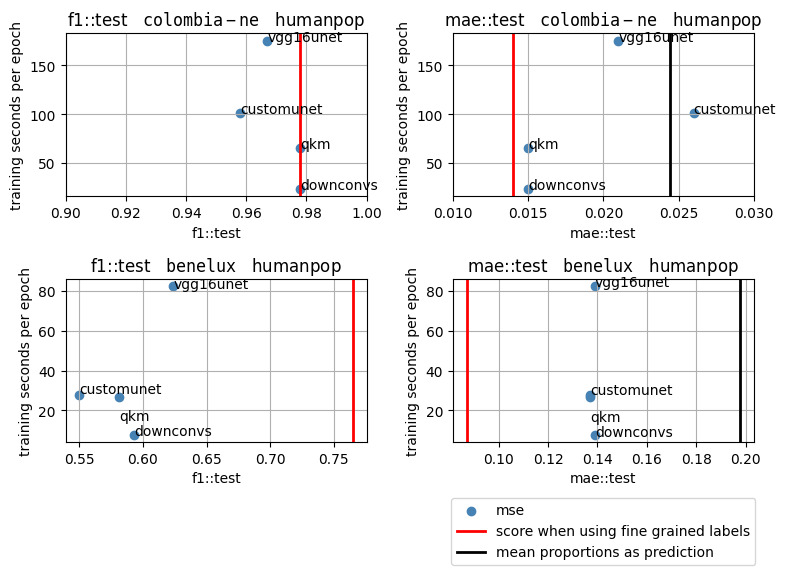}
\label{fig:humanpop-results-vs-time}
\end{figure}

Considering the large difference in models complexity, with these results we believe that \texttt{downconv} and \texttt{qkm} are reasonable choices to address this challenge in the cases considered in this work. Figures \ref{fig:esaworldcover-results-vs-time} and \ref{fig:humanpop-results-vs-time} show how the performance of these simpler models is achieved at a fraction of the training time.  Inference times are around 15\% of train times.

\begin{figure*}[h]
\caption{Predicting label proportions for \texttt{esaworldcover} over \texttt{benelux} classes 1, 2 and 3. White contours represent communes in test and validation. The rest are used in training.}
\centering
\includegraphics[width=1.0\textwidth]{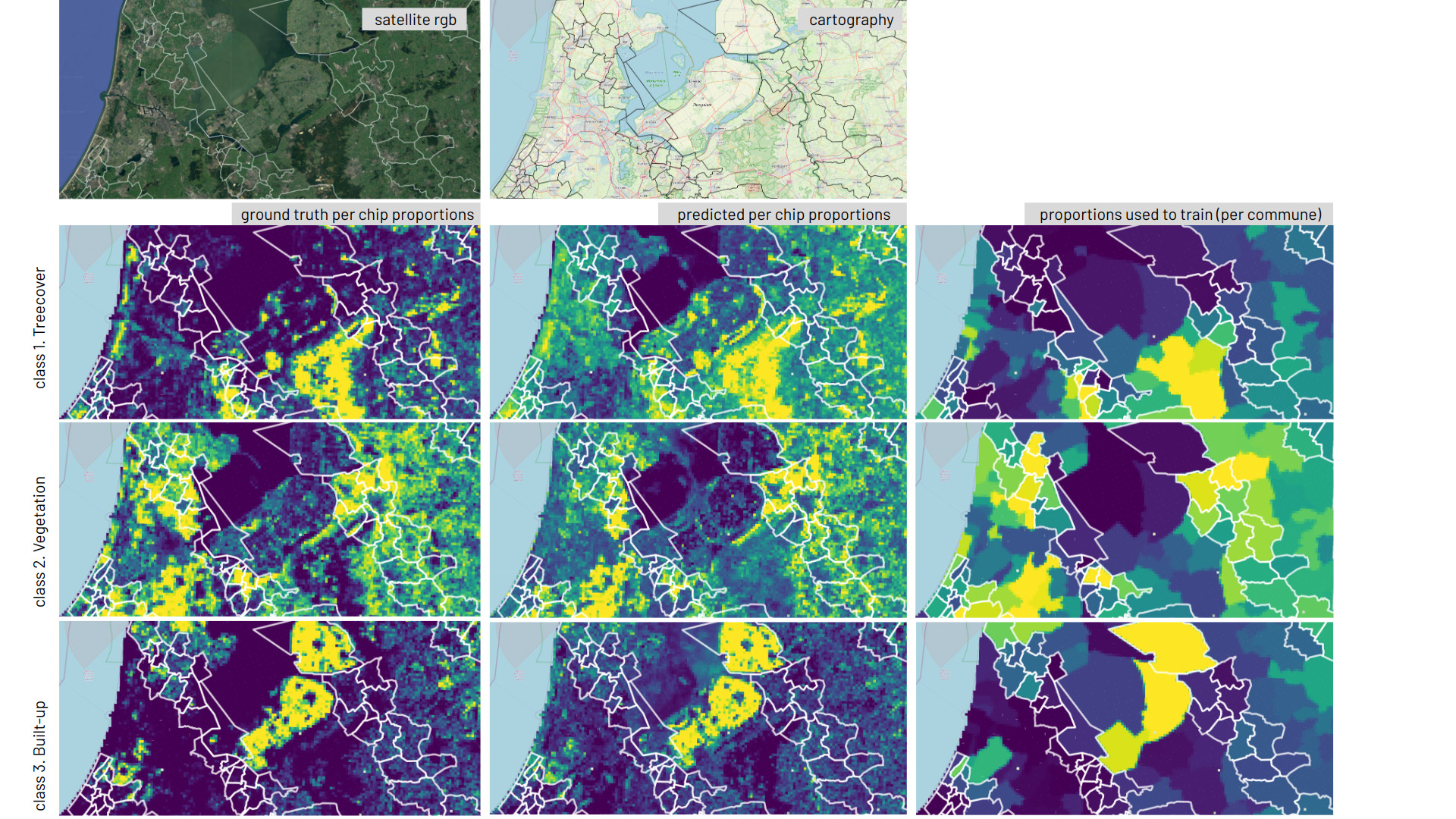}
\label{fig:benelux-esaworldcover-predictions}
\end{figure*}

Moreover, \texttt{downconv} and \texttt{qkm} show no overfitting when training (unlike \texttt{vgg16unet} and \texttt{customunet}), which allows us to put together all predictions in train, val and test on the same map to get an global view of models' performance. See Figure \ref{fig:abstract} in the abstasct and Figures  \ref{fig:benelux-esaworldcover-predictions}, \ref{fig:colombia-esaworldcover-predictions} and \ref{fig:benelux-predictions-humanpop-class2} in the Appendix.

Figures \ref{fig:benelux-esaworldcover-predictions}, \ref{fig:colombia-esaworldcover-predictions} and \ref{fig:benelux-predictions-humanpop-class2} show proportions predictions by the \texttt{downconv} model over large extents of both \texttt{colombia-ne} and \texttt{benelux}.  We include train, validation and test data to make the visualization easier, and mark in white the communes used for validation and train. This model has no overfitting so visualizing all data together should not distort the perception on its performance.

\section{On orbit learning from label proportions}
\label{sec:onorbit}
In the light of the results above and the nature of the LLP problem itself we consider several scenarios for on-orbit machine learning pipelines:

\paragraph{On orbit inference} The fact that models with very few parameters perform reasonably well (see Figures \ref{fig:benelux-esaworldcover-predictions}, \ref{fig:colombia-esaworldcover-predictions} \ref{fig:benelux-predictions-humanpop-class2}) leads to suggest their usage in on-edge devices in general and, specifically in on-orbit platforms. Their small size would also enable to uplink new models updating older ones or for new applications. Besides producing label proportions at the chip level, the models presented in this work also produce segmentation maps which could, in principle, be also generated through inference on orbit.

\paragraph{On orbit training} The necessity of training on orbit is not yet so well established \cite{kothari2020final}, probably \cite{mateo2023orbit} being one of the first works. In any case, with small models, training computing requirements are largely reduced. Furthermore, since we train with label proportions, the actual footprint of the training targets is very small. See Table \ref{tab:llp-sizes} for training targets storage estimation, comparing LLP at communes level with a classical segmentation scenario. Observe the large difference, which might also allow to uplink new training targets to on orbit machine learning training platforms, as training data is processed and curated on the ground.

\begin{table*}
\begin{adjustwidth}{-1cm}{}
\caption{Dataset footprint (storage size) for label proportions at commune level, computed as 2 bytes (for \texttt{float16} to store a number $\in [0,1]$) $\times$ number of classes $\times$ number of communes; and for segmentation labels computed as 1 byte per pixel (for \texttt{uint8} to store a class number) $\times$ (100$\times$100) pixels per chip $\times$ the number of chips (one chip per km$^2$).
\label{tab:llp-sizes}
}
\begin{center}
\begin{tabular}[t]{ccccc}
    \toprule
      &
      \multicolumn{2}{c}{label proportions} & 
      \multicolumn{2}{c}{segmentation labels} \\
       &  {\texttt{benelux}} & \texttt{colombia-ne} & {\texttt{benelux}} & \texttt{colombia-ne} \\
       & {1082 communes} & {244 communes} & {72.2 km$^2$} & {69.2 km$^2$}\\
      \midrule
    \texttt{esaworldcover} &  10.6 Kb  & 2.4 Kb & 705 Mb & 676 Mb \\
    5 classes \\
    \\
    \texttt{humanpop} &  4.2 Kb & 0.9 Kb & 705 Mb & 676 Mb\\
    2 classes \\
    \bottomrule
  \end{tabular}


\end{center}
\end{adjustwidth}
\end{table*}

\paragraph{On orbit volumetry} Taking Sentinel 2 on low earth orbit as an example, a full orbit takes 100 mins and covers a swath 290 Km wide, which represents 290 Km $\times$ 40K Km (Earth's circumference) = 11.6M km$^2$. Since about 29.9\% of the is landmass a full orbit covers in average 3.7M Km$^2$. This represents about 37K Km$^2$/min in a 100 mins orbit, which is roughly the size of our train datasets (see Subsection \ref{sec:data-splitting}).  In our machine, training one epoch (going once through the train dataset) took under 20 secs with \texttt{qkm} or \texttt{downconvs} and inference under 3 secs (see Section \ref{sec:results}). This sheds a dim light of feasibility for on orbit training and inference. However, we are aware that these experiments would have to be performed in platforms akin to the ones on orbit to draw meaningful conclusions.

\section{Conclusion}
\label{sec:conclusions}
In this work we've shown how very small deep learning models (5K parameters) are able to provide a surprising level of fine grained predictions when trained with very coarse label proportions on two tasks (vegetation and human population density prediction). This constitutes a small step in enabling on-orbit machine learning pipelines since we effectively reduce by several orders of magnitude both (1) the on-board computational requirements for training models, and (2) the footprint of the label sets to train on new tasks.

Visual appreciation of the results suggests that this degree of performance could be useful enough for many applications (change detection, estimations, etc.) Furthermore, the results are consistent across two very different world regions in terms of geophysical features and label distributions. A possible interpretation of this behaviour is that the coarse level of the training label proportions acts as a natural regularizer, having a greater effect on the performance of simpler models over more complex ones.

Finally, LLP scenarios based on coarse and tabular data naturally occur in satellite imagery and, they are bound to contribute to compensate the endemic scarcity of fine grained labeled data in EO. However, EO/LLP datasets  are seldom available today and researchers are obliged to manually preprocess existing EO datasets to obtain some label proportions, or integrate them with external data. These ad-hoc approaches hinder comparative studies and, thus, the understanding of the actual reach and applicability of proposed new methods. As a contribution in this direction, in this work we provide four publicly available benchmarking datasets along with an experimental setup.





\bibliography{references} 

\begin{thebibliography}{25}
\providecommand{\natexlab}[1]{#1}
\providecommand{\url}[1]{\texttt{#1}}
\expandafter\ifx\csname urlstyle\endcsname\relax
  \providecommand{\doi}[1]{doi: #1}\else
  \providecommand{\doi}{doi: \begingroup \urlstyle{rm}\Url}\fi

\bibitem[Zhu et~al.(2017)Zhu, Tuia, Mou, Xia, Zhang, Xu, and
  Fraundorfer]{zhu2017deep}
Xiao~Xiang Zhu, Devis Tuia, Lichao Mou, Gui-Song Xia, Liangpei Zhang, Feng Xu,
  and Friedrich Fraundorfer.
\newblock Deep learning in remote sensing: A comprehensive review and list of
  resources.
\newblock \emph{IEEE geoscience and remote sensing magazine}, 5\penalty0
  (4):\penalty0 8--36, 2017.

\bibitem[Agency(2022)]{copernicus2021}
European~Space Agency.
\newblock \emph{Copernicus Sentinel Data Access Annuel Report 2021}.
\newblock ESA, 2022.
\newblock URL
  \url{https://sentinels.copernicus.eu/web/sentinel/-/copernicus-sentinel-data-access-annual-report-2021}.

\bibitem[Selva and Krejci(2012)]{selva2012survey}
Daniel Selva and David Krejci.
\newblock A survey and assessment of the capabilities of cubesats for earth
  observation.
\newblock \emph{Acta Astronautica}, 74:\penalty0 50--68, 2012.

\bibitem[Kothari et~al.(2020)Kothari, Liberis, and Lane]{kothari2020final}
Vivek Kothari, Edgar Liberis, and Nicholas~D Lane.
\newblock The final frontier: Deep learning in space.
\newblock In \emph{Proceedings of the 21st international workshop on mobile
  computing systems and applications}, pages 45--49, 2020.

\bibitem[Ru{\v{z}}i{\v{c}}ka et~al.(2022)Ru{\v{z}}i{\v{c}}ka, Vaughan,
  De~Martini, Fulton, Salvatelli, Bridges, Mateo-Garcia, and
  Zantedeschi]{ruuvzivcka2022ravaen}
V{\'\i}t Ru{\v{z}}i{\v{c}}ka, Anna Vaughan, Daniele De~Martini, James Fulton,
  Valentina Salvatelli, Chris Bridges, Gonzalo Mateo-Garcia, and Valentina
  Zantedeschi.
\newblock Rav{\ae}n: unsupervised change detection of extreme events using ml
  on-board satellites.
\newblock \emph{Scientific reports}, 12\penalty0 (1):\penalty0 16939, 2022.

\bibitem[Mateo-Garcia et~al.(2021)Mateo-Garcia, Veitch-Michaelis, Smith, Oprea,
  Schumann, Gal, Baydin, and Backes]{mateo2021towards}
Gonzalo Mateo-Garcia, Joshua Veitch-Michaelis, Lewis Smith, Silviu~Vlad Oprea,
  Guy Schumann, Yarin Gal, At{\i}l{\i}m~G{\"u}ne{\c{s}} Baydin, and Dietmar
  Backes.
\newblock Towards global flood mapping onboard low cost satellites with machine
  learning.
\newblock \emph{Scientific reports}, 11\penalty0 (1):\penalty0 1--12, 2021.

\bibitem[Mateo-Garc{\'\i}a et~al.(2023)Mateo-Garc{\'\i}a, Veitch-Michaelis,
  Purcell, Longepe, Reid, Anlind, Bruhn, Parr, and Mathieu]{mateo2023orbit}
Gonzalo Mateo-Garc{\'\i}a, Josh Veitch-Michaelis, Cormac Purcell, Nicolas
  Longepe, Simon Reid, Alice Anlind, Fredrik Bruhn, James Parr, and
  Pierre~Philippe Mathieu.
\newblock In-orbit demonstration of a re-trainable machine learning payload for
  processing optical imagery.
\newblock \emph{Scientific Reports}, 13\penalty0 (1):\penalty0 10391, 2023.

\bibitem[Wang et~al.(2022)Wang, Albrecht, Braham, Mou, and Zhu]{wang2022self}
Yi~Wang, Conrad~M Albrecht, Nassim Ait~Ali Braham, Lichao Mou, and Xiao~Xiang
  Zhu.
\newblock Self-supervised learning in remote sensing: A review.
\newblock \emph{arXiv preprint arXiv:2206.13188}, 2022.

\bibitem[Fasana et~al.(2022)Fasana, Pasini, Milani, and
  Fraternali]{fasana2022weakly}
Corrado Fasana, Samuele Pasini, Federico Milani, and Piero Fraternali.
\newblock Weakly supervised object detection for remote sensing images: A
  survey.
\newblock \emph{Remote Sensing}, 14\penalty0 (21):\penalty0 5362, 2022.

\bibitem[Lacoste et~al.(2021)Lacoste, Sherwin, Kerner, Alemohammad,
  L{\"u}tjens, Irvin, Dao, Chang, Gunturkun, Drouin, et~al.]{lacoste2021toward}
Alexandre Lacoste, Evan~David Sherwin, Hannah Kerner, Hamed Alemohammad,
  Bj{\"o}rn L{\"u}tjens, Jeremy Irvin, David Dao, Alex Chang, Mehmet Gunturkun,
  Alexandre Drouin, et~al.
\newblock Toward foundation models for earth monitoring: Proposal for a climate
  change benchmark.
\newblock \emph{arXiv preprint arXiv:2112.00570}, 2021.

\bibitem[Hern{\'a}ndez-Gonz{\'a}lez(2019)]{hernandez2019framework}
Jer{\'o}nimo Hern{\'a}ndez-Gonz{\'a}lez.
\newblock A framework for evaluation in learning from label proportions.
\newblock \emph{Progress in Artificial Intelligence}, 8\penalty0 (3):\penalty0
  359--373, 2019.

\bibitem[Tsai and Lin(2020)]{tsai2020learning}
Kuen-Han Tsai and Hsuan-Tien Lin.
\newblock Learning from label proportions with consistency regularization.
\newblock In \emph{Asian Conference on Machine Learning}, pages 513--528. PMLR,
  2020.

\bibitem[La~Rosa et~al.(2022{\natexlab{a}})La~Rosa, Oliveira, Thiele, Ghamisi,
  and Gloaguen]{la2022weak}
Laura E~Cu{\'e} La~Rosa, D{\'a}rio A~Borges Oliveira, Sam Thiele, Pedram
  Ghamisi, and Richard Gloaguen.
\newblock Weak-supervision based on label proportions for earth observation
  applications from optical and hyperspectral imagery.
\newblock 2022{\natexlab{a}}.

\bibitem[Shi et~al.(2020)Shi, Liu, Wang, Qi, and Tian]{shi2020deep}
Yong Shi, Jiabin Liu, Bo~Wang, Zhiquan Qi, and YingJie Tian.
\newblock Deep learning from label proportions with labeled samples.
\newblock \emph{Neural Networks}, 128:\penalty0 73--81, 2020.

\bibitem[La~Rosa et~al.(2022{\natexlab{b}})La~Rosa, Oliveira, and
  Ghamisi]{la2022learning}
Laura~EC La~Rosa, Dario~AB Oliveira, and Pedram Ghamisi.
\newblock Learning crop type mapping from regional label proportions in
  large-scale sar and optical imagery.
\newblock \emph{arXiv preprint arXiv:2208.11607}, 2022{\natexlab{b}}.

\bibitem[La~Rosa and Oliveira(2022)]{la2022learningCO}
Laura Elena~Cu{\'e} La~Rosa and D{\'a}rio Augusto~Borges Oliveira.
\newblock Learning from label proportions with prototypical contrastive
  clustering.
\newblock In \emph{Proceedings of the AAAI Conference on Artificial
  Intelligence}, volume~36, pages 2153--2161, 2022.

\bibitem[Ding et~al.(2017)Ding, Li, and Yu]{ding2017learning}
Yongke Ding, Yuanxiang Li, and Wenxian Yu.
\newblock Learning from label proportions for sar image classification.
\newblock \emph{Eurasip Journal on Advances in Signal Processing},
  2017:\penalty0 1--12, 2017.

\bibitem[ESA(2023)]{esas2}
ESA.
\newblock Sentinel 2 {MSI Harmonized Level-2A}.
\newblock
  \url{https://developers.google.com/earth-engine/datasets/catalog/COPERNICUS\_S2\_SR\_HARMONIZED},
  2023.

\bibitem[Zanaga et~al.(2021)Zanaga, Van De~Kerchove, De~Keersmaecker,
  Souverijns, Brockmann, Quast, Wevers, Grosu, Paccini, Vergnaud, Cartus,
  Santoro, Fritz, Georgieva, Lesiv, Carter, Herold, Li, Tsendbazar, Ramoino,
  and Arino]{zanaga_daniele_2021_5571936}
Daniele Zanaga, Ruben Van De~Kerchove, Wanda De~Keersmaecker, Niels Souverijns,
  Carsten Brockmann, Ralf Quast, Jan Wevers, Alex Grosu, Audrey Paccini,
  Sylvain Vergnaud, Oliver Cartus, Maurizio Santoro, Steffen Fritz, Ivelina
  Georgieva, Myroslava Lesiv, Sarah Carter, Martin Herold, Linlin Li,
  Nandin-Erdene Tsendbazar, Fabrizio Ramoino, and Olivier Arino.
\newblock Esa worldcover 10 m 2020 v100.
\newblock
  \url{https://developers.google.com/earth-engine/datasets/catalog/ESA_WorldCover_v100},
  October 2021.
\newblock URL
  \url{https://developers.google.com/earth-engine/datasets/catalog/ESA_WorldCover_v100}.

\bibitem[{European Comission JRC}(2016)]{ghsl}
{Columbia University CEISIN} {European Comission JRC}.
\newblock Ghs population grid, derived from gpw4, multitemporal (1975, 1990,
  2000, 2015).
\newblock
  \url{https://developers.google.com/earth-engine/datasets/catalog/JRC\_GHSL\_P2016\_POP\_GPW\_GLOBE\_V1},
  2016.

\bibitem[{European Comission, Eurostat}(2022)]{eurostat_communes}
{European Comission, Eurostat}.
\newblock Administrative and statistica units.
\newblock
  \url{https://ec.europa.eu/eurostat/web/gisco/geodata/reference-data/administrative-units-statistical-units},
  2022.

\bibitem[{Departamento Administrativo Nacional de Estadística,
  Colombia}(2022)]{dane_mgn}
{Departamento Administrativo Nacional de Estadística, Colombia}.
\newblock Marco {G}eoestadístico {N}acional.
\newblock
  \url{https://geoportal.dane.gov.co/geovisores/territorio/mgn-marco-geoestadistico-nacional/},
  2022.

\bibitem[González et~al.(2023)González, Ramos-Pollán, and
  Gallego-Mejia]{gonzalez2023quantum}
Fabio~A. González, Raúl Ramos-Pollán, and Joseph~A. Gallego-Mejia.
\newblock Quantum kernel mixtures for probabilistic deep learning, 2023.

\bibitem[Gonz{\'{a}}lez et~al.(2022)Gonz{\'{a}}lez, Gallego,
  Toledo-Cort{\'{e}}s, and Vargas-Calder{\'{o}}n]{Gonzalez2022LearningFeatures}
Fabio~A. Gonz{\'{a}}lez, Alejandro Gallego, Santiago Toledo-Cort{\'{e}}s, and
  Vladimir Vargas-Calder{\'{o}}n.
\newblock {Learning with density matrices and random features}.
\newblock \emph{Quantum Machine Intelligence}, 4\penalty0 (2), 12 2022.
\newblock ISSN 25244914.
\newblock \doi{10.1007/S42484-022-00079-9}.

\bibitem[Iakubovskii(2019)]{Yakubovskiy:2019}
Pavel Iakubovskii.
\newblock Segmentation models.
\newblock \url{https://github.com/qubvel/segmentation_models}, 2019.

\end{thebibliography}

\clearpage

\section*{Appendix. Additional figures}

\begin{figure*}[h]
\caption{Predicting label proportions for \texttt{esaworldcover} over \texttt{colombia-ne} classes 1, 2 and 3. Recall that class 3 is largely under represented in this dataset (see Figure \ref{fig:class-distribution}). White contours represent communes in test and validation. The rest are used in training.}
\centering
\includegraphics[width=1.0\textwidth]{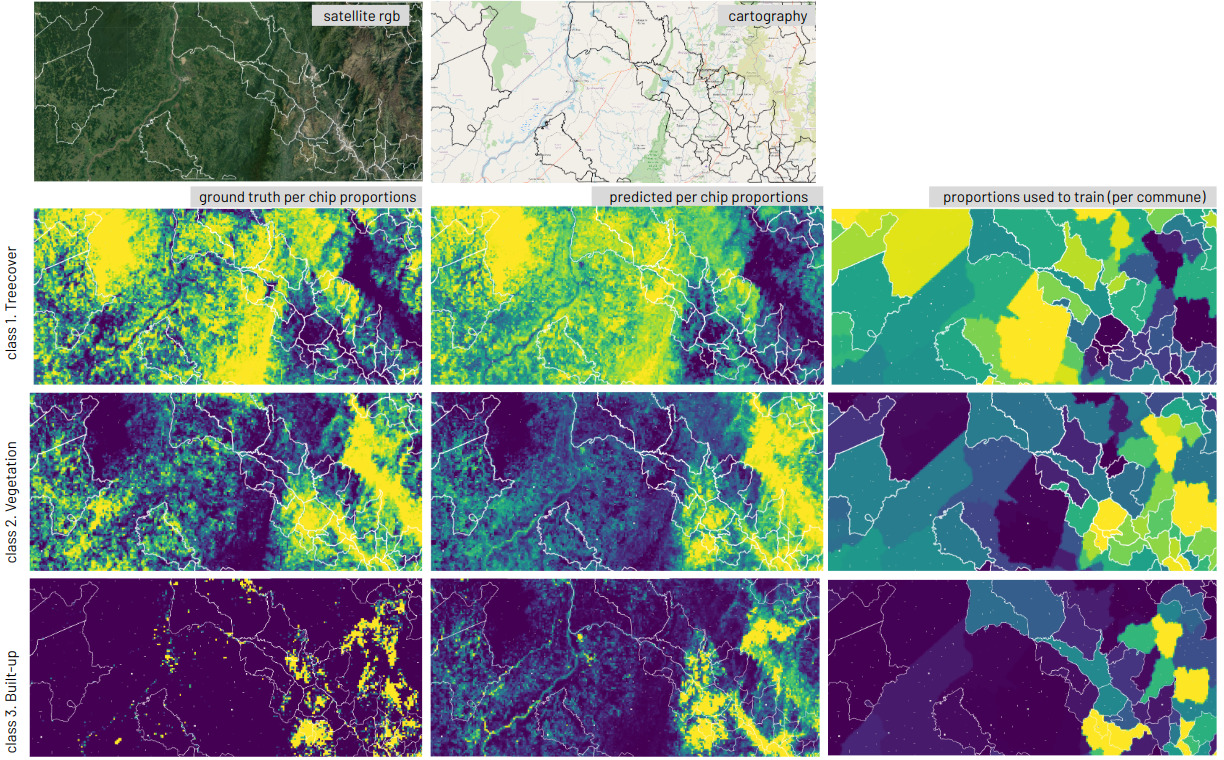}
\label{fig:colombia-esaworldcover-predictions}
\end{figure*}

\begin{figure*}[h]
\caption{Predicting label proportions for \texttt{humanpop} world cover over \texttt{benelux} class 2 (more than 1600 inhabitants/km$^2$). White contours represent communes in test and validation. The rest are used in training.}
\centering
\includegraphics[width=1.0\textwidth]{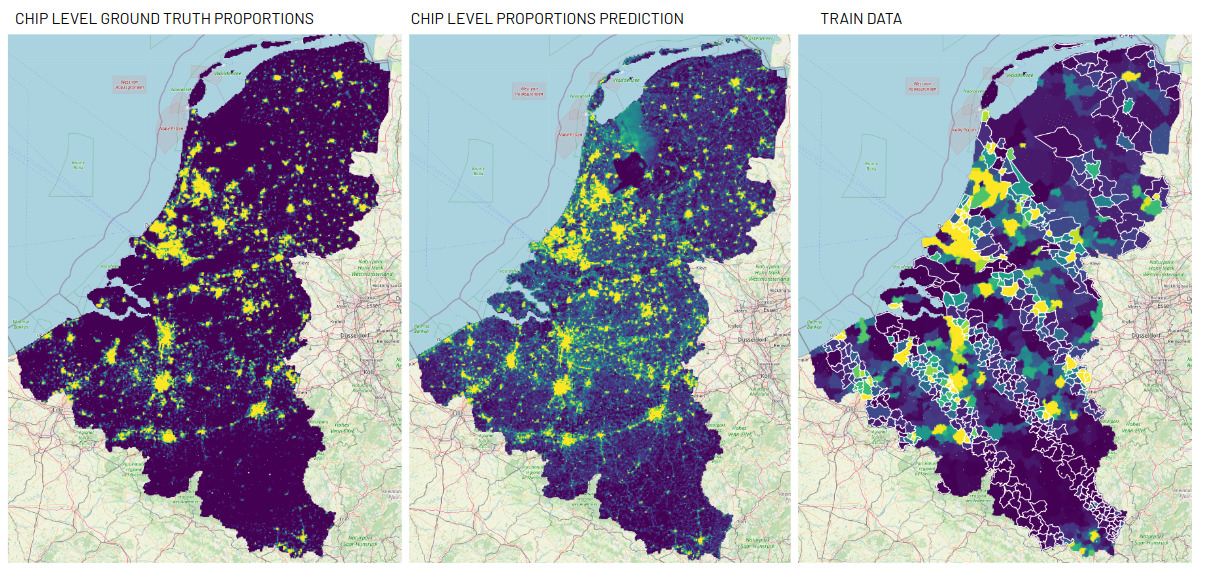}
\label{fig:benelux-predictions-humanpop-class2}
\end{figure*}

\end{document}